\pdfoutput=1

\documentclass[11pt]{article}

\usepackage{EMNLP2022}

\usepackage{times}
\usepackage{latexsym}
\usepackage{amsmath}
\usepackage{graphicx}
\usepackage{amsfonts}
\usepackage{float}
\usepackage{pgfplots}
\usepackage{booktabs}
\usepackage{multicol}
\usepackage{multirow}
\usepackage{arydshln} 
\usepackage{mathtools}
\usepackage{txfonts}
\usepackage{pgfplots}
\usepackage{subfigure}
\usepackage{times}
\usepackage{latexsym}
\usepackage{bm}
\usepackage{pgfplots}

\usepackage[T1]{fontenc}

\usepackage{makecell}

\usepackage[utf8]{inputenc}

\usepackage{microtype}

\usepackage{inconsolata}
\usepackage{enumitem}
\usepackage{ulem}

\newcommand\blfootnote[1]{%
  \begingroup
  \renewcommand\thefootnote{}\footnote{#1}%
  \addtocounter{footnote}{-1}%
  \endgroup
}
\usepackage{float}

\title{Task Compass: Scaling Multi-task Pre-training with Task Prefix}

\author{Zhuosheng Zhang$^{1*}$, Shuohang Wang$^{2}$, Yichong Xu$^{2}$, Yuwei Fang$^{2}$, \\\bf Wenhao Yu$^{3*}$, Yang Liu$^{2}$, Hai Zhao$^{1}$, Chenguang Zhu$^{2}$ and Michael Zeng$^{2}$ \\
$^1$Shanghai Jiao Tong University, Shanghai, China \\
$^2$Microsoft Cognitive Services Research, Redmond, WA, USA \\ 
$^3$University of Notre Dame, Notre Dame, IN, USA \\
{$^1$\tt zhangzs@sjtu.edu.cn, zhaohai@cs.sjtu.edu.cn};\\ 
{$^2$\tt \{shuowa, yicxu, yuwfan, yaliu10, chezhu, nzeng\}@microsoft.com};
{$^3$\tt wyu1@nd.edu}
}

\begin{document}
\maketitle

\blfootnote{* Work done when Zhuosheng Zhang and Wenhao Yu interned at Microsoft Cognitive Services Research group. This work was partially supported by Key Projects of National Natural Science Foundation of China (U1836222 and 61733011).}

\begin{abstract}
Leveraging task-aware annotated data as supervised signals to assist with self-supervised learning on large-scale unlabeled data has become a new trend in pre-training language models. Existing studies show that multi-task learning with large-scale supervised tasks suffers from negative effects across tasks. To tackle the challenge, we propose a task prefix guided multi-task pre-training framework to explore the relationships among tasks. We conduct extensive experiments on 40 datasets, which show that our model can not only serve as the strong foundation backbone for a wide range of tasks but also be feasible as a probing tool for analyzing task relationships. The task relationships reflected by the prefixes align transfer learning performance between tasks. They also suggest directions for data augmentation with complementary tasks, which help our model achieve human-parity results on commonsense reasoning leaderboards. Code is available at \url{https://github.com/cooelf/CompassMTL}

\end{abstract}

\section{Introduction}

Recent years have witnessed a growing interest in leveraging a unified pre-trained language model (PrLM) to solve a wide range of natural language processing tasks \cite{tay2022unifying,chowdhery2022palm,xie2022unifiedskg,zhang2022survey}. The pre-training recipe of a PrLM is driving from self-supervised learning \cite{peters2018deep,radford2018improving,devlin2019bert,lan2019albert,clark2019electra} to multi-task learning (MTL) with a mixture of standard self-supervised tasks and various supervised tasks, which takes advantage of learning from both large-scale unlabeled corpus and high-quality human-labeled datasets \cite{ExploringTL2019Raffel,aribandi2021ext5}.\footnote{Since multi-task pre-training is often implemented as an additional large-scale learning stage between language model pre-training and fine-tuning, it is also known as multi-task pre-fine-tuning in literature \cite{aghajanyan2021muppet}.} Benefitting from supervision from related tasks, MTL approaches reduce the cost of curating deep learning models for an individual task and provide a shared representation that is generally applicable for a range of tasks \cite{wu2020understanding}.

\begin{figure}
	\centering
	\includegraphics[width=0.48\textwidth]{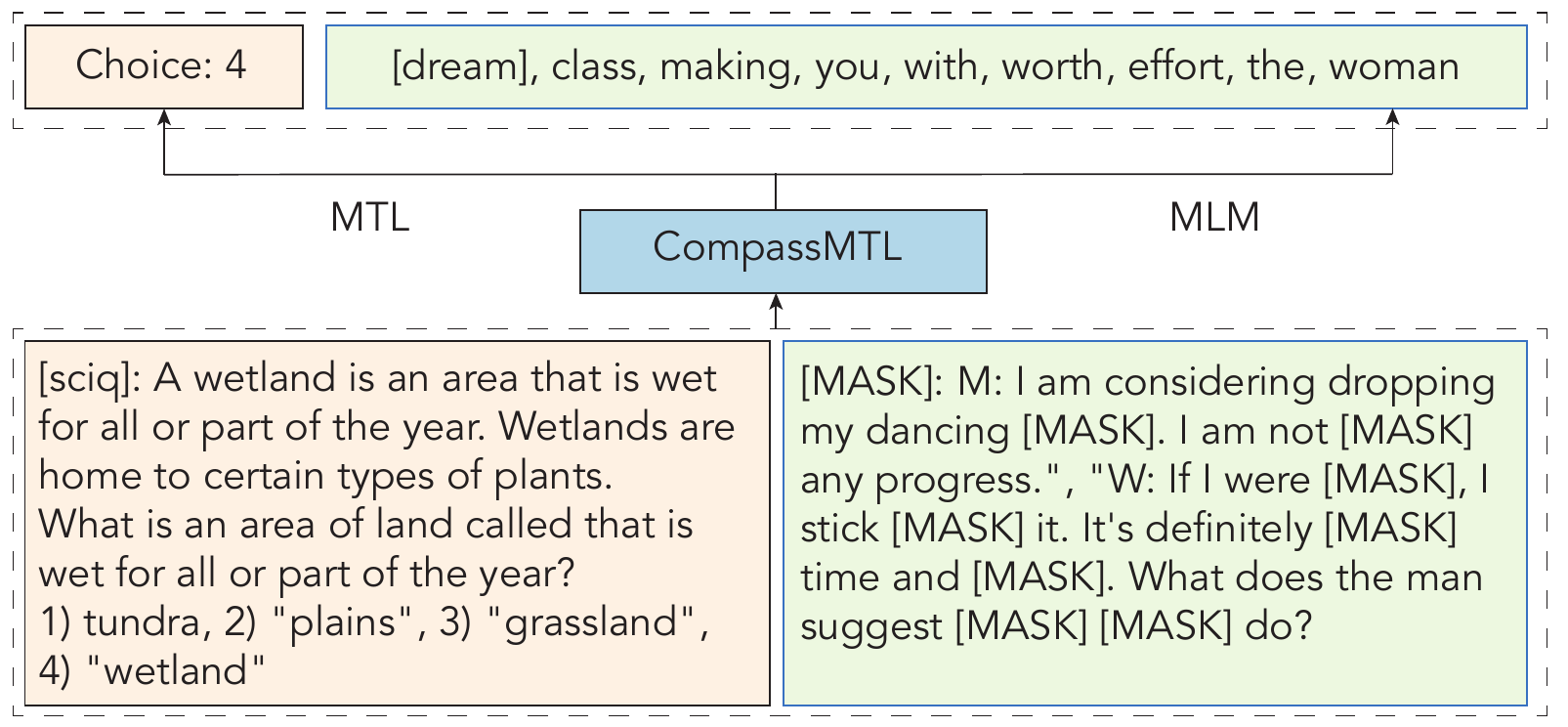}
	\caption{Input-output view. We append a task prefix for each data sequence to capture common patterns from the dataset and require the model to predict some randomly masked prefixes to capture task differences.
	}
	\label{fig:compass}
\end{figure}

In the research line of multi-task learning for PrLMs, a typical solution is to cast all tasks into a text-to-text format and utilize an encoder-decoder PrLM such as T5 to predict the target sequences \cite{ExploringTL2019Raffel,aribandi2021ext5}. Despite the extensive efforts on leveraging supervised tasks in strengthening PrLMs, the latest trend is extreme scaling of task numbers, with little attention paid to the relationships between tasks \cite{sanh2021multitask,wei2021finetuned}. \citet{aribandi2021ext5} investigated co-training transfer effects amongst task-families and empirically found that tasks in different families may have side effects between each other, e.g., summarization tasks generally seem to hurt performance on other task families such as dialogue system \cite{mehri2020dialoglue}, natural language inference \cite{bowman2015large}, and commonsense reasoning \cite{lourie2021unicorn}.

When the task number scales up, the training of PrLMs would be more vulnerable to negative transfer due to the severe inconsistency of domain and data distribution between tasks \cite{wu2020understanding,padmakumar-etal-2022-exploring}. As one of the key concepts underlying MTL, task relationships potentially provide an effective basis for employing PrLMs in a more effective and interpretable way.

To handle the issue of negative transfer during multi-task learning, early studies have taken task relationships into account by employing a dual-process model architecture that is composed of a shared encoder and task-specific layers. The two parts are supposed to integrate the common features of all the learning tasks and explore the task relationship in a predefined manner \cite{zheng2019metadata,liu2019multi,bai2020multi,ma2021adaptive}, respectively. However, these methods require additional modifications to model architecture and increase the model complexity and computation cost. Therefore, they are suboptimal for applying to PrLMs in terms of generality and computational bottlenecks.

All the considerations above lay down our goal to investigate simple yet effective ways to measure the task relationship without additional cost and keep the generality of PrLMs. In this work, we propose a prefix-guided multi-task learning framework (CompassMTL) to explore the mutual effects between tasks (Figure \ref{fig:compass}) and improve model performance with complementary tasks. Targeting natural language understanding (NLU) tasks, we employ a discriminative PrLM\footnote{Also known as encoder-only PrLMs. As this work focuses on NLU tasks, we find that encoder-only PrLMs are competitive based on our empirical studies though they may lose generalizability on natural language generation tasks.} as the backbone model and train the model on 40 tasks. Experimental results show that our model achieves human-parity performance on commonsense reasoning tasks. We further probe into the task relationship entailed in the tasks prefix representations, finding that the measured relationship highly correlates with task-to-task transfer performance, and it is also of referenced value for optimizing the PrLM on a target task with its complementary tasks during MTL, i.e., fewer tasks with better performance.

In summary, our contributions are three folds:

1) A unified discriminative multi-task PrLM for NLU tasks will be released as a strong counterpart for the dominant T5-based encoder-decoder PrLMs trained with MTL.

2) A probing tool of using task prefix to explore the task relationships in large-scale MTL. We observe that the task relationships reflected by the prefixes manifest a correlation with transfer learning performance, and they help our model achieve better results with complementary tasks.

3) State-of-the-art results on a variety of NLU tasks, especially human-parity benchmark performance on commonsense reasoning leaderboards, i.e., HellaSwag and $\alpha$NLI.







\begin{figure*}[t]
	\centering
	\includegraphics[width=0.95\textwidth]{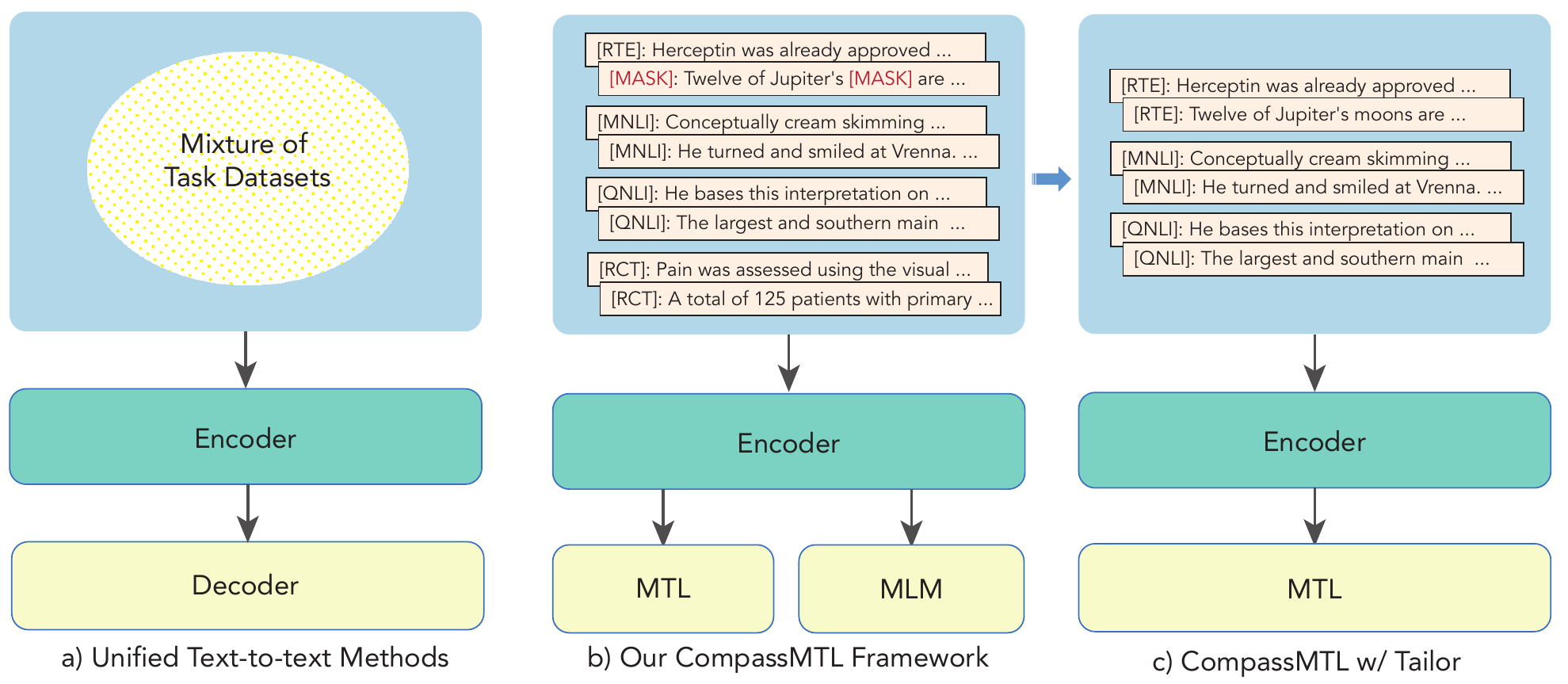}
	\caption{Comparison with existing paradigms of multi-task learning. Typical unified text-to-text methods include T5 \cite{ExploringTL2019Raffel}, ExT5 \cite{aribandi2021ext5}, FLAN \cite{wei2021finetuned}, and T0 \cite{sanh2021multitask}.}
	\vspace{-0.1in}
	\label{fig:framework}
\end{figure*}

\section{Background and Related Work}
\subsection{Self-supervised Pre-training}
PrLMs are commonly pre-trained on large-scale corpora and then used for fine-tuning individual tasks. One of the most widely-used pre-training tasks is masked language modeling (MLM), which first masks out some tokens from the input sentences and then trains the model to predict them by the rest tokens. There are derivatives of MLM including permuted language modeling in XLNet \cite{yang2019xlnet} and sequence-to-sequence MLM in MASS \cite{MASSMS2019Song} and T5 \cite{ExploringTL2019Raffel}. Beyond the general-purpose pre-training, domain-adaptive pre-training and task-adaptive pre-training have attracted attention in recent studies. 



1) Domain-adaptive Pre-training. To incorporate specific in-domain knowledge, domain-aware pre-training is designed, which directly post-trains the original PrLMs using the domain-specific corpus. Popular models have been proposed in the dialogue domain \cite{whang2020effective,wu2020tod}, as well as in the medical and science domains \cite{BioBERTAP2020Lee, SciBERTAP2019Beltagy, ClinicalBERTMC2019Huang,yu2022dict}.

2) Task-adaptive Pre-training. The goal of task-adaptive pre-training is to capture task-specific skills by devising the pre-training tasks. The popular application scenarios include logical reasoning and dialogue-related tasks \citet{kumar2020deep,gu2020dialogbert,zhang-zhao-2021-structural,li2021deep}. For example, \citet{whang2021response} proposed various utterance manipulation strategies, including utterance insertion, deletion, and retrieval, to maintain dialog coherence. 


\subsection{Multi-task Learning for PrLMs}
Our concerned MTL in the field of PrLMs is partially related to the studies of task-adaptive pre-training discussed above. The major difference is that the PrLMs in MTL are fed with human-annotated datasets instead of those automatically constructed ones for self-supervised tasks. Figure \ref{fig:framework} overviews the paradigms of MTL PrLMs. 
Existing methods in this research line mostly vary in model architectures and training stages. For example, MT-DNN \cite{liu2019multi} applied multi-task learning to train a shared model on all the target datasets in the fine-tuning stage, and there are several task-aware output modules to adapt the shared representations to each task. 
Recent studies, such as ExT5 \cite{aribandi2021ext5}, T0 \cite{sanh2021multitask}, and FLAN \cite{wei2021finetuned}, commonly applied an Encoder-Decoder architecture and convert a variety of tasks into the same text-to-text format and train those tasks jointly (Figure \ref{fig:framework}-a).  We argue that they are not the optimal solution considering the model complexity and the gap between original and transformed task formats, especially for natural language understanding tasks that are in a discriminative manner, e.g., classification, multiple-choice, etc. Actually, there are studies \cite{mccann2018natural,keskar2019unifying,li2019unified,khashabi-etal-2020-unifiedqa} that transform traditional tasks into other formats like reading comprehension or question answering and achieve better results than prior methods. These studies motivate us to explore superior model backbones and data formats, especially for the application in NLU tasks.

\subsection{Modeling Task Relationships in MTL}
Modeling task relationships is a classic topic in deep learning studies.
\citet{bingel-sogaard-2017-identifying} studied the research question about what task relations make gains in traditional natural language processing tasks and investigated when and why MTL works in sequence labeling tasks such as chunking, sentence compression, POS tagging, keyphrase detection, etc.
\citet{wu2020understanding} found that task data alignment can significantly affect the performance of MTL and proposed architecture with a shared module for all tasks and a separate output module for each task. 


Since these methods require additional modifications of model architecture, they are suboptimal for employment in PrLMs, considering computational bottlenecks and generality when task scaling. In the era of pre-trained models, \citet{geva2021s} analyzed the behavior transfer in PrLMs between related jointly-trained tasks such as QA and summarization and thus provided evidence for the extrapolation of skills as a consequence of multi-task training. ExT5 \cite{aribandi2021ext5} evaluated the transfer performance among task families in a multi-task co-training setup and observed that negative transfer is common, especially when training across task families. Although there are recent studies that insert prompts to describe the task requirements in the data sequences \cite{liu2021pre,su2022transferability,qin2021exploring,vu2022spot}, it is still not clear whether the prompts help negative transfer or whether the prompts necessarily capture task relationships. In this work, we find that using task prefixes along with the MLM for prefix prediction effectively indicates task relationships and helps MTL with fewer datasets but better performance.


\begin{figure*}[t]
	\centering
	\includegraphics[width=0.96\textwidth]{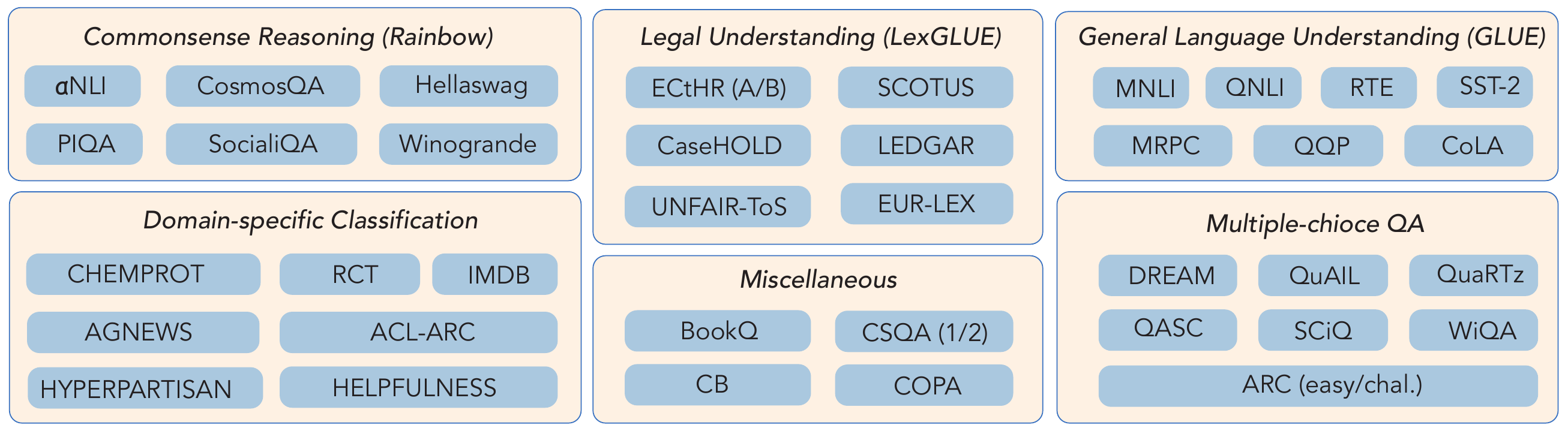}
	\caption{Task taxonomy used in this work.}
	\label{fig:tasks}
	\vspace{-0.1in}
\end{figure*}

\section{Methodology}
\subsection{Task Format}\label{sec:format}
According to prior studies \cite{mccann2018natural,keskar2019unifying,khashabi-etal-2020-unifiedqa}, the benchmark results on a task can be affected dramatically by training a model on different formats of the same dataset. In contrast to converting all tasks in a text-to-text manner, we choose to model our tasks in a multiple-choice-like format to minimize the format transformation for NLU tasks. Our transformation aims to ensure that each example in a task has a specific number of $k$ candidate options during the multi-task training stage. The original pair-wise input texts are regarded as context and question in the view of the multiple-choice problem. If there is only one text given, then the question will be kept empty. For the outliers, the data will be processed as follows (Examples are provided in Appendix \ref{appendix:data}).

1) If the number of candidate options > $k$, the redundant options will be randomly discarded;

2) If the number of candidate options < $k$, add  "N/A" placeholder options.

3) If the ground truth is a list, randomly select a correct option from the gold list and randomly sample $k-1$ negative options from the held-out set\footnote{The held-out set is composed of all the candidate items in each gold list.} except the left items in the gold list.

4) If the ground truth is a list and there is an empty choice, construct the truth option manually. For example, "there is no violation"; the negative examples are constructed as the same as 3).

As a result, each training example will be formed as a sequence like \{[Prefix]: context, question, option\}, where [Prefix] indicates the task name in natural language such as [hellawag] prepended to each data example.

\subsection{CompassMTL}
Our model is encoder-only, which is based on the DeBERTa architecture \cite{he2021debertav3}. The model is trained by using both the supervised task objective and the standard self-supervised denoising objective as described below. 

Suppose that we have a dataset $\mathcal{D} = \{(y_i, c_i, q_i, r)\}^N_{i=1}$, where $c_i$ represents the context, $q_i$ represents the question, $r$ denotes a set of answer options $r = \{r_1, \dots, r_k\}$, and $y_i$ is the label. $N$ is the number of training data. Each data example is formed as $x = $ \textup{[CLS]} \textup{[Prefix]} $c_i$ \textup{[SEP]} $q_i$ $r_j$ \textup{[SEP]},\footnote{The task prefixes are added to the model vocabulary as additional tokens to avoid tokenization.} $r_j \in r$. The goal is to learn a discriminator $g(\cdot,\cdot)$ from $\mathcal{D}$. For the supervised task, the loss function is:
$
   \mathcal{L}_{mtl} = - \sum_{i=1}^N\sum_{j=1}^k \log(g(c_i, q_i \circ  r_j)).
$

At the inference phase, given any new context $c_i$, question $q_i$ and options $r$, we use the discriminator to calculate $g(c_i, q_i \circ  r_j)$ as their matching score where $\circ$ denotes concatenation. The option with the highest score is chosen as the answer for the $i$-th example.

Let $\hat{x}_{i}$ denote the masked sequence where a certain proportion of tokens in $x_{i}$ are randomly replaced with a special [MASK] symbol. Using $\hat{x}_{i}$ as the input fed to the model in parallel with $x$, the self-supervised denoising objective is computed in the way of MLM:
$
    \mathcal{L}_{mlm} = -\sum_{i=1}^N\sum_{j\in \mathcal{M}} \log\: p_{\theta}(t_{i,j}\mid \hat{x}_i),
$
where $t_{i,j}$ is the $j$-th token in $x_i$ and $\mathcal{M}$ denotes the index set of masked tokens for which the loss will be computed. 
To encourage the model to learn from both supervised and self-supervised signals, we combine $\mathcal{L}_{mtl}$ and $\mathcal{L}_{mlm}$ during training:
$
    \mathcal{L} = \mathcal{L}_{mtl} + \lambda \mathcal{L}_{mlm}
$
where $\lambda$ is a hyper-parameter to balance the weight of the training objectives.

Compared with traditional MTL methods, CompassMTL is data-centric, without any modification of model architecture (Figure \ref{fig:framework}-b). It can be regarded as an efficient implementation of the traditional MTL method composed of a shared representation module and multiple task-aware modules. Since the data from the same datasets share the same task prefix, the prefix is supposed to reflect the common patterns from the dataset, which works in a similar operational principle to the shared representation module. During the training with our self-supervised objective, task prefixes will be randomly masked in a specific probability.\footnote{Each token in the input sequence will be masked in the same probability, including the task prefix and the rest tokens.}
The model is required to distinguish the task prefixes and predict the right prefix according to the input data. Therefore, the task differences will also be necessarily captured.

\subsection{Task Relationship Exploration}
Regarding the task prefixes as the compass to navigate the task relationships, it is possible to use our  framework to analyze the relevance of tasks (Section \ref{sec:prob}). Our model for prefix probing experiments is slightly revised from CompassMTL, which only uses the MLM objective and is fed by the data without options to alleviate possible shortcuts in options. After the model is pre-trained with MTL, we fetch the prefix embeddings from the model embedding layer and calculate the Pearson correlation between each task pair with min-max normalization. Assuming that we have $n$ tasks, the process will result in $n\times n$ correlation scores to indicate the task relationships. 

For a target task, we can directly rank the top-related tasks according to the correclation scores and use those complementary tasks for MTL before fine-tuning a target task (Figure \ref{fig:framework}-c).

\section{Experiments}
\subsection{Datasets}
There are 40 datasets used for training our multi-task model, some of which are collected from GLUE \citep{wang-etal-2018-glue}, SuperGLUE \cite{wang2019superglue}, Rainbow \cite{lourie2021unicorn}, and LexGLUE \cite{chalkidis2021lexglue}. Figure \ref{fig:tasks} illustrates the composition of our task families.

\paragraph{GLUE}
GLUE (The General Language Understanding Evaluation benchmark) \citep{wang-etal-2018-glue} is a collection of 9 various tasks for sentence-level classification. We only use 8 of them: CoLA\citep{warstadt2018neural}, SST-2 \citep{socher2013recursive}, MRPC \citep{dolan2005automatically}, STS-B \citep{cer2017semeval}, QQP \citep{chen2018quora},  QNLI \citep{Rajpurkar2016SQuAD}, MNLI \citep{nangia2017repeval} and RTE \citep{bentivogli2009fifth}. 


\paragraph{Rainbow} Rainbow \cite{lourie2021unicorn} is a suite of commonsense question answering tasks including $\alpha$NLI \cite{bhagavatula2019abductive}, CosmosQA \cite{huang2019cosmos}, HellaSwag \cite{zellers2019hellaswag}, PIQA \cite{bisk2020piqa}, SocialIQA \cite{sap2019social}, Winogrande \cite{sakaguchi2020winogrande}.

\paragraph{LexGLUE} LexGLUE (Legal General Language Understanding Evaluation) \cite{chalkidis2021lexglue} is a collection of datasets for evaluating model performance across a diverse set of legal NLU tasks, which contain 7 subtasks, namely ECtHR (Task A), ECtHR (Task B), SCOTUS, EUR-LEX, LEDGAR, UNFAIR-ToS, and CaseHOLD.

\paragraph{Domain-specific Classification} We use seven datasets that cover specific domains (biomedical and computer science publications, news, and reviews) following \citet{gururangan-etal-2020-dont}. The datasets are CHEMPROT \cite{kringelum2016chemprot}, RCT \cite{dernoncourt2017pubmed}, ACL-ARC \cite{jurgens2018measuring}, HYPERPARTISAN \cite{kiesel2019semeval}, AGNEWS \cite{zhang2015character}, HELPFULNESS \cite{mcauley2015image}, and IMDB \cite{maas2011learning}.

\paragraph{Multiple-choice QA} The datasets include DREAM \cite{sun2019dream}, QuAIL \cite{rogers2020getting}, QuaRTz \cite{tafjord2019quartz}, WiQA \cite{tandon2019wiqa}, QASC \cite{khot2020qasc}, SCiQ \cite{welbl2017crowdsourcing}, ARC \cite{clark2018think}. We follow \citet{sanh2021multitask} to organize this task family. 

\paragraph{Miscellaneous} The other datasets are BookQ \cite{clark2019boolq}, CB \cite{de2019commitmentbank}, CommonsenseQA v1/v2 \cite{talmor2019commonsenseqa,talmor2021commonsenseqa}, and COPA \cite{roemmele2011choice}. BoolQ, CB, and COPA are also collected in SuperGLUE \cite{wang2019superglue}. We select those tasks as they can be easily transformed into our unified format.

\begin{table*}[t]
    \centering
    \small
    \setlength{\tabcolsep}{1.8pt}
    {
    \centering
        {
    \begin{tabular}{l|ccc|cccccc|c}
    \toprule
     \bf Model  & \bf Arch. & \bf Tasks & \bf Params.  & \bf $\alpha$NLI & \bf CosmosQA & \bf HellaSwag & \bf PIQA & \bf SocialIQA & \bf Winogrande  & \bf Average \\
     \midrule
     UNICORN  & Enc-Dec & 6 & 770M & 79.5	& 83.2 &	83.0 &	82.2 &	75.5 & 	78.7 &	80.4\\ 
       ExT5   & Enc-Dec & 107 & 770M &  82.3	& 85.9	& 89.0	& 85.0 & 	79.7 &	82.5	& 84.1\\
       \midrule
       ExDeBERTa & Enc only & 40 & 567M & 87.9 & 85.3& 83.6 & 85.5 & 79.6 & 87.0 & 84.8 \\
     CompassMTL & Enc only & 40& 567M & 91.7  & 87.8 & 95.6 &  87.3 & 81.7 & 89.6 & 89.0 \\
     \quad w/ Tailor & Enc only & 14 & 567M & \textbf{92.5} & \textbf{88.8} & \textbf{96.1} & \textbf{88.3} & \textbf{82.2} & \textbf{90.5} & \bf 89.7 \\
         \bottomrule
    \end{tabular}}}
    \caption{Results on the Rainbow commonsense reasoning  validation sets. The baseline models are UNICORN$_{\texttt{large}}$ \cite{lourie2021unicorn} and ExT5$_{\texttt{large}}$ \cite{aribandi2021ext5}. ExDeBERTa is our imitation of ExT5-style \cite{aribandi2021ext5} MTL training by using DeBERTa backbone trained on 40 datasets with a multi-task objective of self-supervised denoising and supervised task objective, after which is transferred to each individual task. "w/ Tailor" denotes multi-task training with related datasets (14-subset) according to our discovery in Section \ref{sec:aug}.}
    \label{tab:rainbow}
\end{table*}

\begin{table*}[t]
    \centering
    \small
    \setlength{\tabcolsep}{4.2pt}
    {
    \begin{tabular}{lcccccccccccccc}
      \toprule
         \multirow{2}{*}{\bf Method}  & \multicolumn{2}{c}{\bf ECtHR (A)} & \multicolumn{2}{c}{\bf ECtHR (B)} & \multicolumn{2}{c}{\bf SCOTUS} & \multicolumn{2}{c}{\bf EUR-LEX} & \multicolumn{2}{c}{\bf LEDGAR}  & \multicolumn{2}{c}{\bf UNFAIR-ToS} & \bf CaseHOLD\\
         & $\mathrm{\muup}$-$\mathrm{F_1}$ & $\mathrm{m}$-$\mathrm{F_1}$ & $\mathrm{\muup}$-$\mathrm{F_1}$ & $\mathrm{m}$-$\mathrm{F_1}$ & $\mathrm{\muup}$-$\mathrm{F_1}$ & $\mathrm{m}$-$\mathrm{F_1}$ & $\mathrm{\muup}$-$\mathrm{F_1}$ & $\mathrm{m}$-$\mathrm{F_1}$ & $\mathrm{\muup}$-$\mathrm{F_1}$ & $\mathrm{m}$-$\mathrm{F_1}$ & $\mathrm{\muup}$-$\mathrm{F_1}$ & $\mathrm{m}$-$\mathrm{F_1}$ & $\mathrm{\muup}$/$\mathrm{m}$-$\mathrm{F_1}$ \\
         \midrule
BERT            & 71.2 & 63.6 & 79.7 & 73.4 & 68.3 & 58.3 & 71.4 & 57.2 & 87.6 & 81.8 & 95.6 & 81.3 & 70.8 \\
RoBERTa         & 69.2 & 59.0 & 77.3 & 68.9 & 71.6 & 62.0 & 71.9 & \bf 57.9 & 87.9 & 82.3 & 95.2 & 79.2 & 71.4 \\
DeBERTa         & 70.0 & 60.8 & 78.8 & 71.0 & 71.1 & 62.7 & \bf 72.1 & 57.4 & 88.2 & 83.1 & 95.5 & 80.3 & 72.6 \\
Longformer      & 69.9 & \bf 64.7 & 79.4 & 71.7 & 72.9 & 64.0 & 71.6 & 57.7 & 88.2 & 83.0 & 95.5 & 80.9 & 71.9 \\
BigBird         & 70.0 & 62.9 & 78.8 & 70.9 & 72.8 & 62.0 & 71.5 & 56.8 & 87.8 & 82.6 & 95.7 & 81.3 & 70.8 \\
Legal-BERT      & 70.0 & 64.0 & 80.4 & \bf 74.7 & 76.4 & 66.5 & \bf 72.1 & 57.4 & 88.2 & 83.0 & 96.0 & 83.0 & 75.3 \\
CaseLaw-BERT    & 69.8 & 62.9 & 78.8 & 70.3 & 76.6 & 65.9 & 70.7 & 56.6 & \bf 88.3 & 83.0 & 96.0 & 82.3 & 75.4 \\
\midrule
ExDeBERTa & - & - & - &  - & - & - & - & - & - & - & - & - & 74.8 \\
CompassMTL & 71.7 & 60.7 & 80.6 & 73.2 & \bf 77.7 & \bf 68.9 & 67.2 & 42.1 & 88.1 & 82.3 & \bf 96.3 & \bf 84.3 & 76.1\\
\quad w/ Tailor & \bf 73.0 & \bf 64.7& \bf 80.7 & 72.3 & 76.3 & 68.6 & 66.9 & 44.9 & \bf 88.3 & \bf 83.2 & 96.2 & 83.2 & \bf  78.1\\
\bottomrule
    \end{tabular}
    }
    \caption{Results on LexGLUE test sets. The baseline results except ours in the last column are from \citet{chalkidis2021lexglue}. Since the LexGlue tasks except CaseHold are multi-label classification problems, the ExDeBERTa model is not directly applicable for those tasks without extra task-specific fine-tuning; thus, the results are not reported. "w/ Tailor" denotes multi-task training with the seven datasets in the same LexGLUE family.
    }
    \label{tab:lex_glue}
    \vspace{-0.15in}
\end{table*}

\subsection{Implementations}
Our model is implemented using Pytorch and based on the Transformers Library \cite{wolf2019huggingface}. To save computation, we initialize our model with the released checkpoints of DeBERTa-V3-Large, and the hyper-parameter setting generally follows DeBERTa \cite{he2021debertav3}. Our experiments are run on 8x32GB Tesla A100 GPUs. The maximum input sequence length is 512. Similar to \citet{lourie2021unicorn}, the implementation of CompassMTL includes two procedures. We first conduct multi-task pre-training on all the datasets and then continue to train on each target dataset alone to verify the performance. For multi-task pre-training, we use a peak learning rate of 6e-6 with a warm-up rate of 0.1. We run up to 6 epochs using a batch size of 128. The masking ratio of MLM is 0.25, and $\lambda$ is set to 0.1. To avoid large-scale datasets dominating the pre-training, the training data is randomly sampled by a limit of 10$k$ on the maximum dataset size according to \citet{ExploringTL2019Raffel}. For fine-tuning experiments, the initial learning rate is selected in \{3e-6, 6e-6, 8e-5\} with a warm-up rate of 0.1. The batch size is selected in \{16, 32\}. The maximum number of epochs is chosen from \{6,10\}. More fine-tuning details are available in Appendix \ref{appendix:ft}.

\subsection{Main Results}
Our main results are reported on the Rainbow and LexGLUE benchmark datasets for comparisons with public methods. As the statistics shown in Tables \ref{tab:rainbow}-\ref{tab:lex_glue}, we see that CompassMTL models outperform the related public models in general. Specifically, it is observed that our encoder-only models yield better performance than the T5-based encoder-decoder models under similar model sizes. Further, the comparison in the second column discloses the potential to achieve comparable or better performance by multi-task learning with related tasks (w/ Tailor). How to find the related tasks and use them to enhance model performance will be discussed in the following section.

\begin{figure*}
	\centering
	\includegraphics[width=0.96\textwidth]{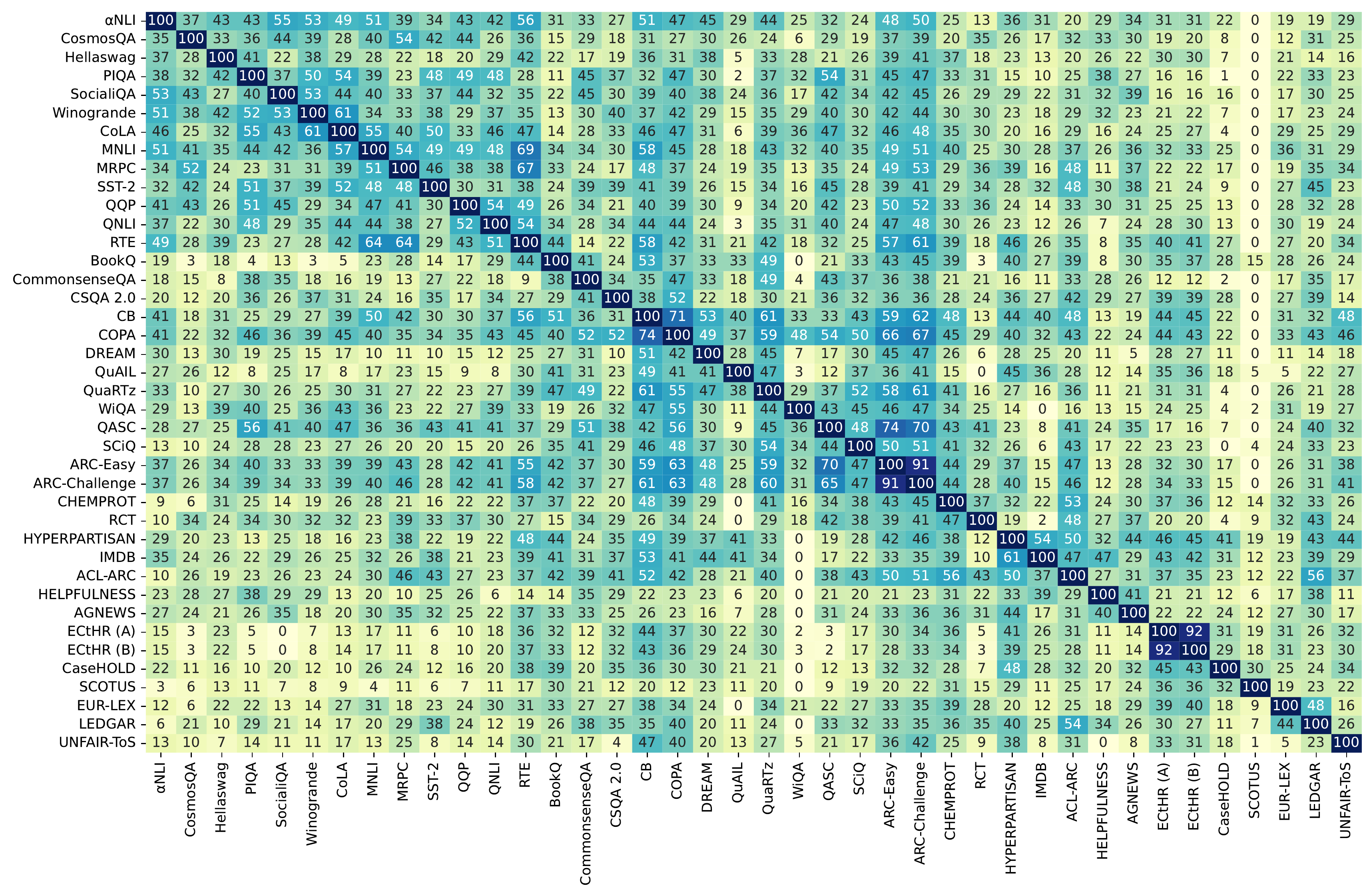}
	\caption{Heatmap of task relationships probed by prefix embeddings.}
	\label{fig:probing}
\end{figure*}


\section{Analysis}

\subsection{Ablation Study}\label{appendix:abl}
Table \ref{tab:abl} presents our ablation study to dive into the effectiveness of different training objectives and the influence of task prefixes in our method. For the training objectives, MTL and MLM denote the training objectives of $\mathcal{L}_{mtl}$ and $\mathcal{L}_{mlm}$, respectively. The results suggest that both supervised and self-supervised tasks contribute to the overall model performance, and the supervised task is more beneficial than the self-supervised task in our study. Further, to inspect the role of the task prefixes, we ablate the model with three conditions: 1) \textit{must}: the prefixes are masked with the probability of 1.0; 2) \textit{no}: the prefixes are masked with the probability of 0.0; 3) \textit{only}: only prefixes will be masked, i.e., the prefix of each example will be masked, while the other tokens are left as original.\footnote{Note that if we ideally mask all the prefixes, the prefix tokens will not appear in the input sequence; thus, the prefix embeddings will not be updated. To avoid this issue, we follow the standard practice of training BERT-like models, where the masked tokens will experience extra processes: 1) 80\% of the time, we replace masked input tokens with mask symbols; 2) 10\% of the time, we replace masked input tokens with a random word; 3) The rest of the time (10\% of the time) we keep the masked input tokens unchanged.} The results in Table \ref{tab:abl} show that using prefixes (Prefix$_{\texttt{must}}$ and Prefix$_{\texttt{only}}$) indeed boosts the model performance generally. 

\begin{table}[t]
    \centering
    \small
    \setlength{\tabcolsep}{16pt} 
    {
    \centering
        {
    \begin{tabular}{lc}
    \toprule
     \bf Model  & \bf Accuracy \\
     \midrule
     Single & 84.6 \\
     \midrule 
     CompassMTL & 89.4  \\
     \quad - MTL & 85.0\\
     \quad - MLM & 88.8\\
     \midrule 
    Prefix$_{\texttt{must}}$ & 89.3\\
    Prefix$_{\texttt{no}}$ & 88.9\\
    Prefix$_{\texttt{only}}$ & 89.1\\
         \bottomrule
    \end{tabular}}}
    \caption{Ablation Study of the training objectives and task prefixes. We calculate the average accuracy scores on the development sets of all the 40 datasets.}
    \label{tab:abl}
\end{table}

\subsection{Relationship Probing}\label{sec:prob}
Figure \ref{fig:probing} illustrates the heatmap of task relationships probed by prefix embeddings. We see that the datasets inside the same task family (e.g., GLUE and Rainbow) correlate highly with each other. The LexGLUE tasks are less related to other tasks because the texts are mainly legal descriptions. In addition, the correlation scores also accord with the common practice of data augmentation. For example, the NLI datasets (MNLI, QNLI, RTE) share close relevance, and it is helpful to initialize parameters from an MNLI model to fine-tune RTE \cite{liu2019roberta,qu2020coda}.

We are interested in whether the probed relationship scores coordinate with the model performance transferred between tasks. We first obtain transfer accuracy between tasks in a dual-task training setup \cite{aribandi2021ext5}. Assume that we have 13 source tasks from GLUE and Rainbow tasks and 5 target tasks ($\alpha$NLI, HellaSwag, MRPC, PIQA, QNLI, and RTE). We first train individual models using the mixture of training sets from each pair of source and target tasks, and then evaluate the model on the validation set of the target dataset. 
As a result, we have $5\times 13$ transfer results. For each target dataset, we calculate Pearson correlation between relationship scores and transfer accuracy among the source datasets. In Table \ref{ana:correlation}, we find that the relationship scores are positively bound up with the transfer performance. The results indicate the potential to find related tasks by the relationship scores. In other words, the relationship scores essentially reflect task relationships.

Task relationships may also be reflected by shallow token distributions, such as vocabulary overlap or sentence length. To investigate if our relationship probing can be replaced by comparing the token distributions, we further analyze the correlation between the similarity of token distributions and dual-task transfer accuracy. For sentence length, we first calculate the absolute values of the average length difference between source and target datasets and then convert them to negative values (intuitively less difference in length, more close the relationship). The vocab overlap of the source and target datasets is also computed for comparison. The similarity between datasets reflects weak correlations with the transfer accuracy (2/5 and 3/5 datasets, respectively in Table \ref{ana:correlation}). These results are less consistent than our probing method, which indicates that our method mines more complex patterns toward task relationships.


\begin{table}[t]
    \centering \small
\setlength{\tabcolsep}{3pt} 
\begin{tabular}{lcccccc}
    \toprule
    \textbf{Dataset} & \textbf{RTE} & \textbf{MRPC}  & \textbf{QNLI} & \textbf{HellaSwag}  &\textbf{$\alpha$NLI} & \textbf{Avg.} \\
    \midrule
    Probing & 0.19 & 0.22 & 0.38 & 0.12 & 0.51 & 0.28 \\
Length & -0.12 & 0.43 & -0.17 & 0.04 & -0.07 & 0.02 \\
Vocab & 0.37 & -0.27 & -0.001 & 0.09 & 0.31 & 0.10 \\
    \bottomrule
  \end{tabular}
  \caption{Pearson correlation between each relationship measure and the transfer accuracy. \label{ana:correlation}}
\end{table}

\begin{table}[t]
    \centering \small
\setlength{\tabcolsep}{2pt} 
\begin{tabular}{l|c|cc|ccc}
    \toprule
    \textbf{Model} & \bf Tasks & \textbf{RTE} & \textbf{MRPC} & \textbf{QNLI} & \textbf{HellaSwag} & \textbf{$\alpha$NLI} \\
    \midrule
    Single & 1 &61.4  & 89.2  & 95.0 &  95.1 & 91.3\\
    40-fullset& 40 & \textbf{92.8}  & 90.4 & 95.5 &  95.6 &91.7  \\
    \midrule
    Top 5 & 5 & 92.4 & \textbf{91.9} & 95.3 &  95.6  & 91.6\\
    Family & 6/7 & 91.4 &  90.2 & 95.0 & 95.7  & 91.9  \\
    14-subset  & 14 & 91.8 & 90.3  & \textbf{95.6} & \textbf{96.1} & \textbf{92.5} \\
    \bottomrule
  \end{tabular}
  \caption{Complementary transfer results using different mixtures of datasets for MTL. The last three rows represent the mixture in different granularity inspired by our relationship probing. 
}\label{ana:aug}
\end{table}

\subsection{Complementary Transfer} \label{sec:aug}
To inspect whether using more datasets always leads to better performance and whether using the most related datasets can lead to competitive results. In this part, we conduct a complementary transfer analysis by selecting a group of datasets to train an MTL model and fine-tuning the model on target datasets. Four choices of dataset mixture are compared: 1) 40-fullset: the same as our basic setting of CompassMTL in this work; 2) Top-5 ranked dataset according to based on our probed relationship scores; 3) Family: the datasets belonged to the same family with the target dataset, i.e., 6 datasets for Rainbow tasks and 7 datasets for GLUE tasks; 4) 14-subset: the mixture of Rainbow and GLUE datasets.

Table \ref{ana:aug} presents the comparison results. We observe that the top-5 ranked variant yields comparable, even better results than the others, which indicates that models trained with more datasets may not always bring benefits. The results also indicate that small-scale datasets (e.g., MRPC and RTE), which have relatively high average correlation scores with the other datasets, are more likely to benefit from the complementary transfer. With the tasks scaling up, the performance (family $\rightarrow$ 14-subset) may improve as more related tasks are involved in training.

\begin{table}[t]
    \centering
\setlength{\tabcolsep}{6.3pt} \small
\begin{tabular}{lcccccc}
    \toprule
    \multirow{2}{*}{\textbf{Model}} & \multicolumn{2}{c}{\textbf{SQuADv1.1}} & \multicolumn{2}{c}{\textbf{SQuADv2.0}} & \textbf{NER} \\
     & \textbf{EM} & \textbf{F1} & \textbf{EM} & \textbf{F1} & \textbf{F1} \\
    \midrule
    Baseline & 88.8 & 94.8 & 87.1  & 90.5 & 96.5 \\
    CompassMTL  & 89.7 & 95.1 & 88.5 & 91.3 & 96.9 \\
    \bottomrule
  \end{tabular}
  \caption{Results on the SQuAD v1.1/V2.0 and CoNLL2003 (NER) development sets. The evaluation metrics are Exact-Match (EM) and F1 scores.}\label{exp-format}
\end{table}

\begin{table}[t]
    \centering \small
\setlength{\tabcolsep}{12pt} 
\begin{tabular}{lcccccc}
    \toprule
    \textbf{Model} & \textbf{HellaSwag} & \textbf{$\alpha$NLI} \\
    \midrule
    Human Performance  & 95.60 & 92.90\\
    \midrule
    Previous SOTA & 94.87 & 92.20 \\
    Our Results  & 95.94 & 92.80 \\
    \bottomrule
  \end{tabular}
  \caption{Leaderboard tests of HellaSwag and $\alpha$NLI.}\label{ana:human}
\end{table}

\begin{table*}[htb]
    \centering
    \small
    \setlength{\tabcolsep}{8.8pt}
    {
    \centering
        {
    \begin{tabular}{lccccccc}
    \toprule
     \bf Model  & \bf $\alpha$NLI & \bf CosmosQA & \bf HellaSwag & \bf PIQA & \bf SocialIQA & \bf Winogrande  & \bf Average \\
     \midrule
     T5 & 68.5 & 69.6 & 56.6 & 67.7 & 65.1 & 62.4 & 65.0 \\
     UNICORN & 65.3 & 72.8 & 56.2 & 73.3 & 66.1 & 61.8 & 65.9 \\
     CompassMTL & \textbf{69.1} & \textbf{72.6} & \textbf{57.7} & \textbf{73.6} & \textbf{66.6} & \textbf{64.9} & \textbf{67.4}\\
         \bottomrule
    \end{tabular}}}
    \caption{Results on the Rainbow validation sets by using T5-base as the backbone model.}
    \label{tab:t5}
\end{table*}

\subsection{Human-parity on Commonsense Reasoning Leaderboards}
Table \ref{ana:human} presents our test evaluation on the official leaderboards of HellaSwag\footnote{\url{https://leaderboard.allenai.org/hellaswag/submissions/public}} and  $\alpha$NLI\footnote{\url{https://leaderboard.allenai.org/anli/submissions/public}}. The submissions are based on the ensemble of three models selected according to Section \ref{sec:aug}. Compared with public methods that use much larger PrLMs, model ensemble, and knowledge graphs, our models establish new state-of-the-art results and reach human-parity performance.

\subsection{Beyond The Unified Format}
To verify whether our model can be employed for tasks that are unavailable to be transformed into our unified format, we evaluate the effectiveness of CompassMTL by using the typical reading comprehension datasets SQuAD v1.1/2.0 \cite{Rajpurkar2016SQuAD,rajpurkar2018know} and named entity recognition (NER) dataset CoNLL 2003 \cite{tjong2003introduction}, which represent extractive question answering and sequence labeling task formats, respectively. We first replicate the baselines for fine-tuning QA and NER tasks using the Transformers toolkit.\footnote{\url{https://github.com/huggingface/transformers}.} For comparison, we initialize the baseline parameters with our model weights to see if CompassMTL is better than the baselines. Results in Table \ref{exp-format} show that our model is generally effective across formats. The results also indicate that CompassMTL can serve as a strong off-the-shelf representation encoder that is applicable for new tasks without needing to be pre-trained again.

\subsection{Implementation Using The T5 Backbone}\label{appendix:t5}
Although our method is implemented by the encoder-only backbone to compete in NLU tasks, it is supposed to be generally applicable to other kinds of PrLMs, such as encoder-decoder T5. To verify the effectiveness, we employ the pre-trained T5-base model \cite{ExploringTL2019Raffel} as the backbone. We use the Rainbow datasets for MTL and convert the data into text-to-text format following the standard processing for T5 training, with task prefixes inserted before each data sequence. The baselines are the single-task T5 trained on each individual task and UNICORN \cite{lourie2021unicorn} trained on the Rainbow datasets. Results in Table \ref{tab:t5} verify that our method is generally effective.

\section{Conclusions}
This work presents a task prefix guided multi-task method by making use of task prefix to explore the mutual effects between tasks and improve model performance with complementary tasks. Our released model can not only serve as the strong foundation backbone for a wide range of NLU tasks but also be used as a probing tool for analyzing task relationships. Our model shows generalizable advances over tasks in diverse formats and establishes human-parity results on commonsense reasoning tasks. Based on our pre-trained model, we find that the prefixes necessarily reflect task relationships, which correlate with transfer learning performance between tasks and suggest directions for data augmentation of complementary tasks. In summary, our work has the following prospects for future studies:

\noindent\textbf{1) Collaborative multi-task learning of PrLMs.} The recipe of using task prefixes in conjunction with prefix prediction in MLM training has shown effective for large-scale MTL pre-training.

\noindent\textbf{2) Suggestive choice for data augmentation.} The task relationships probed by the prefix embeddings have shown informative in finding the complementary tasks. Using complementary tasks helps obtain better performance for a target task, especially for small-scale task datasets.

\noindent\textbf{3) Guidance for skill-aware model evaluation.} The discovery of task relationships may help determine redundant datasets that assess similar patterns of models. Recently, there has been a trend to evaluate the comprehensive skills of deep learning models by using a large number of datasets \cite{srivastava2022beyond}, the selection of distinctive datasets can be guided by our relationship discovery criteria to avoid evaluation redundancy and save computation.

\noindent{\textbf{Limitations.}}
We acknowledge the major limitation of this work is that our model may not readily apply to new tasks. It is based on the common assumption of MTL that the set of tasks is known at training time. Adaptation to new tasks could be future work.


\normalem
\bibliography{anthology,custom}
\bibliographystyle{acl_natbib}

\appendix

\begin{table*}[htb]
    \small
    {
    \centering
        {
    \begin{tabular}{p{7.2cm}p{3cm}p{4.6cm}}
    \toprule
     \bf Context  & \bf Question & \bf Option(s) \\
     \midrule
    {}[sciq] A wetland is an area that is wet for all or part of the year. Wetlands are home to certain types of plants. & What is an area of land called that is wet for all or part of the year? & ["tundra", "plains", "grassland", "wetland"] \\
\midrule
{}[commonsense\_qa] revolving door & A revolving door is convenient for two direction travel, but it also serves as a security measure at a what? & [ "bank", "library", "department store", "mall", \sout{"new york"}] \\
\midrule
{}[dream] M: I am considering dropping my dancing class. I am not making any progress.", "W: If I were you, I stick with it. It's definitely worth time and effort. & What does the man suggest the woman do? & [ "Consult her dancing teacher.", "Take a more interesting class.", "Continue her dancing class.", "N/A"] \\
\midrule
{}[scotus] The Interstate Commerce Commission, acting under § 19a of the Interstate Commerce Act, ordered the appellant to furnish certain inventories, schedules, maps and charts of its pipe line property ... & - & ["Unions", "Economic Activity", "Judicial Power", "Federalism"]\\
\midrule
{}[unfair\_tos] you must provide accurate and complete data during the registration and update your registration data if it changes . & - & ["there is no unfair contractual term", "Limitation of liability", "Unilateral termination", "Arbitration"] \\
        \bottomrule
    \end{tabular}}}
    \caption{Examples of transformed datasets.}
    \label{tab:data}
\end{table*}
\section{Appendix}

\subsection{Examples of transformed datasets}\label{appendix:data}
Table \ref{tab:data} shows examples of transformed datasets. The first column presents the standard multiple-choice dataset, followed by four types of outlier datasets (Section \ref{sec:format}) that are transformed into our unified format. 

\subsection{Fine-tuning Details}\label{appendix:ft}
According to Section \ref{sec:format}, our training datasets are converted into a multiple-choice-like format for multi-task pre-training. During fine-tuning, because our evaluated GLUE and Rainbow tasks for public comparisons are either single-label classification or multiple-choice tasks, the conversion would not affect the performance according to our preliminary experiments as the predictions can be easily mapped to the original formats by choosing the best-ranked options. For the other tasks, such as the multi-label classification tasks in LexGLUE, where the conversion will result in the clip of ground-true labels, we use the original datasets for fine-tuning and initialize the corresponding baseline models with our pre-trained weights after MTL. The criteria for choosing the baseline models for different types of tasks basically follows the standard practice in literature \cite{he2021debertav3,chalkidis2021lexglue}.

\end{document}